\documentclass{article}

\usepackage{arxiv}

\usepackage[utf8]{inputenc}    
\usepackage[T1]{fontenc}       
\usepackage{hyperref}          
\usepackage{url}               
\usepackage{booktabs}          
\usepackage{amsfonts}          
\usepackage{amsmath, amssymb}  
\usepackage{nicefrac}          
\usepackage{microtype}         
\usepackage{graphicx}          
\usepackage{natbib}            
\usepackage{doi}               
\usepackage{xcolor}            
\usepackage{tcolorbox}         
\usepackage{multirow}          
\usepackage{subcaption}        
\usepackage{enumitem}          
\usepackage{bookmark}          
\usepackage{cleveref}          

\definecolor{accentred}{HTML}{C0392B}
\definecolor{accentblue}{HTML}{2C5F8D}
\definecolor{accentgreen}{HTML}{1E8449}
\definecolor{accentorange}{HTML}{D35400}
\definecolor{accentpurple}{HTML}{6C3483}
\definecolor{lightgray}{HTML}{F2F3F4}
\definecolor{codebg}{HTML}{F8F9F9}

\tcbuselibrary{skins, breakable}

\newtcolorbox{takeaway}[1][]{
  enhanced, breakable,
  colback=lightgray, colframe=accentorange,
  fonttitle=\bfseries,
  title={#1},
  boxrule=0.8pt,
  arc=2pt, left=6pt, right=6pt, top=4pt, bottom=4pt,
  attach boxed title to top left={xshift=6pt, yshift=-3pt},
  boxed title style={colback=accentorange, sharp corners, boxrule=0pt}
}

\newtcolorbox{definitionbox}[1][]{
  enhanced, breakable,
  colback=lightgray, colframe=accentblue,
  fonttitle=\bfseries,
  title={#1},
  boxrule=0.8pt,
  arc=2pt, left=6pt, right=6pt, top=4pt, bottom=4pt,
  attach boxed title to top left={xshift=6pt, yshift=-3pt},
  boxed title style={colback=accentblue, sharp corners, boxrule=0pt}
}

\newtcolorbox{methodbox}[1][]{
  enhanced, breakable,
  colback=white, colframe=accentgreen!70!black,
  fonttitle=\bfseries,
  title={#1},
  boxrule=0.6pt,
  arc=2pt, left=6pt, right=6pt, top=4pt, bottom=4pt,
  attach boxed title to top left={xshift=6pt, yshift=-3pt},
  boxed title style={colback=accentgreen!70!black, sharp corners, boxrule=0pt}
}

\newtcolorbox{highlightbox}[1][]{
  enhanced, breakable,
  colback=lightgray, colframe=accentpurple,
  fonttitle=\bfseries,
  title={#1},
  boxrule=0.8pt,
  arc=2pt, left=6pt, right=6pt, top=4pt, bottom=4pt,
  attach boxed title to top left={xshift=6pt, yshift=-3pt},
  boxed title style={colback=accentpurple, sharp corners, boxrule=0pt}
}

\newcommand{\method}[1]{\textsf{\small #1}}

\usepackage{titlesec}
\titleformat{\section}{\large\bfseries\scshape}{\thesection.}{0.6em}{}
\titleformat{\subsection}{\normalsize\bfseries}{\thesubsection}{0.6em}{}
\titleformat{\subsubsection}{\normalsize\bfseries\itshape}{\thesubsubsection}{0.6em}{}

\title{Beyond Tokens: A Unified Framework for\\Latent Communication in LLM-based\\Multi-Agent Systems}

\author{
 \textbf{Yingzhuo Liu}\\
  \textit{Companion repository:} \url{https://github.com/enochliu98/Awesome-Latent-Communication}\\
  \texttt{\small liuyingzhuo86@bupt.edu.cn}
}

\date{}

\renewcommand{\shorttitle}{Latent Communication}

\hypersetup{
  pdftitle={Beyond Tokens: A Unified Framework for Latent Communication in LLM-based Multi-Agent Systems},
  pdfsubject={cs.MA, cs.CL, cs.AI},
  pdfauthor={Anonymous Authors},
  pdfkeywords={Latent Communication, Multi-Agent LLMs, KV-Cache, Hidden States, Embeddings, Agent Communication, Survey},
  colorlinks=true,
  linkcolor=accentblue,
  citecolor=accentblue,
  urlcolor=accentred
}

\begin{document}
\maketitle

\begin{abstract}
\noindent
Multi-agent systems built on large language models (LLMs) typically communicate through natural-language messages. Although readable and interoperable, this protocol requires the sender to decode internal computation into tokens and the receiver to encode those tokens again; under finite message budgets, it may also omit uncertainty and alternative hypotheses. \emph{Latent communication} instead transfers continuous model state---embeddings, hidden states, or key--value (KV) caches---directly between agents. This survey provides a scoped review of this emerging area through searches of arXiv, ACL Anthology, and OpenReview, supplemented by backward snowballing, with a cut-off date of 15 July 2026. We organise eighteen representative works using three axes: \textbf{WHAT} is transferred, \textbf{WHICH} sender and receiver spaces or layers are aligned, and \textbf{HOW} the received state is fused. Beyond taxonomy, we derive deployment-oriented design rules, separate semantic latent messaging from systems-level cache reuse, and identify the conditions under which reported efficiency gains are comparable. The synthesis exposes recurring trade-offs among information capacity, transport cost, architectural dependence, trainability, and auditability. It also motivates a research agenda spanning cross-architecture alignment, secure latent channels, compression, evaluation standardisation, and hybrid text--latent protocols.
\end{abstract}

\keywords{Latent Communication \and Multi-Agent LLMs \and KV-Cache \and Hidden States \and Embeddings \and Agent Communication \and Survey}

\section{Introduction}
\label{sec:intro}

Multi-agent systems built on top of large language models (LLMs) have rapidly become a workhorse for complex reasoning, planning, code generation, scientific question answering, and tool orchestration~\citep{wu2023autogen, hong2023metagpt, li2023camel, liu2026rainbowarena}. In the canonical architecture, several specialised LLM agents --- each typically instantiated as a separate model call with its own role prompt --- collaborate by exchanging \emph{natural language} messages. A planner proposes a strategy in text; a critic reads the proposal and replies in text; a coder edits the plan in text; and so on. The result is a visible, inspectable, human-readable communication trace that doubles as an audit log and a debugging surface. The way such a system partitions a complex task across agents --- \emph{which} subtask to assign to \emph{which} agent --- is itself a non-trivial design choice, and recent work has begun to study adaptive task-decomposition strategies empirically~\citep{liu2025std}.

Despite its success, the \emph{text-only} communication protocol is being increasingly questioned. Three structural limitations stand out:

\begin{enumerate}[leftmargin=*, itemsep=2pt]
  \item \textbf{Inference cost.} Every message forces the sender to autoregressively decode a token sequence and the receiver to prefill that sequence. The exact cost depends on attention implementation, context length, batching, and cache reuse, but it grows with both message length and the number of communication rounds.
  \item \textbf{Potential information loss during discretisation.} A text message maps a rich internal state to a discrete sequence. Per token, the selected symbol conveys at most $\log_2 V$ bits under a fixed vocabulary; a multi-token message can, of course, encode more. The practical concern is that finite message budgets and decoding choices may omit uncertainty, alternatives, and task-relevant detail.
  \item \textbf{Redundancy and ambiguity of natural language.} Generated text is optimised for linguistic fluency rather than task-relevant information density. Idioms, hedging, and vague referents add overhead; disagreements about role assignment or background knowledge can render entire messages irrecoverable.
\end{enumerate}

In response, a new line of work---collectively called \emph{latent communication}---lets agents exchange continuous internal representations directly: embeddings, hidden states, or key--value (KV) caches. Skipping part of the text round trip can preserve task-relevant state or reduce inference time under favourable conditions. It also introduces costs: the channel is opaque to humans, architecture-dependent, and harder to inspect, secure, and version.

The field has grown explosively. The accompanying repository \emph{Awesome-Latent-Communication} already tracks more than fifteen distinct methods, and the diversity of design choices is striking: some methods transmit embeddings, others transmit hidden states, still others transmit KV-caches. Some methods align the last layer of the sender to the first layer of the receiver; others align all layers. Some fuse information by concatenation; others by prepending, addition, or learned cross-attention. Some are training-free; others require distillation. A new researcher entering the area is therefore confronted by a fragmented landscape with no shared vocabulary.

\paragraph{Contributions.} This paper turns a fragmented set of methods into a comparable design space. Specifically:

\begin{itemize}[leftmargin=*, itemsep=2pt]
  \item We define the scope of \emph{latent communication} and document a reproducible review protocol, including search sources, inclusion criteria, boundary cases, and a fixed cut-off date.
  \item We propose a 3-axis decomposition---\textbf{WHAT} (transmitted representation), \textbf{WHICH} (sender--receiver alignment), and \textbf{HOW} (receiver-side fusion)---and use it to compare eighteen representative works.
  \item We derive deployment-oriented design rules that connect representation choice to model heterogeneity, context length, bandwidth, training budget, and auditability.
  \item We identify evaluation confounders and formulate an actionable research agenda for alignment, security, compression, theory, and deployment.
\end{itemize}

\paragraph{Organisation.} \Cref{sec:methodology} describes the review protocol and scope; \Cref{sec:prelim} introduces notation; \Cref{sec:case} develops the motivation and boundary conditions; \Cref{sec:framework} presents the WHAT / WHICH / HOW framework and design guidance; \Cref{sec:training} analyses implementation and training regimes; \Cref{sec:open} formulates the research agenda; and \Cref{sec:related} distinguishes adjacent areas. Detailed method cards and the cross-paper benchmark synthesis are provided in \Cref{sec:methods,sec:benchmarks}.

\section{Review Scope and Methodology}
\label{sec:methodology}

\subsection{Search Strategy and Cut-off}

We searched arXiv, ACL Anthology, and OpenReview using combinations of \emph{latent communication}, \emph{activation communication}, \emph{hidden-state communication}, \emph{KV-cache sharing}, \emph{cache transfer}, \emph{multi-agent LLM}, and \emph{language-model agents}. We supplemented keyword search with backward snowballing from the retrieved papers and periodic screening of the companion repository. The review covers work available on or before \textbf{15 July 2026}. Because terminology is not yet standardised, the search deliberately combines representation-centric terms (e.g., hidden state) with systems-centric terms (e.g., cache transfer).

\subsection{Eligibility and Boundary Cases}

A work enters the primary corpus when it satisfies three conditions: (1) at least two model instances, roles, or agent modules exchange information; (2) the exchanged object includes a continuous internal representation or reusable inference state; and (3) that object affects the receiver's computation without requiring a complete text-only decode--encode round trip. We include peer-reviewed papers and technically substantive preprints because much of the area post-dates established venue cycles. We exclude ordinary prompt sharing, semantic retrieval caches that store only text or external embeddings, and single-agent latent reasoning unless they clarify the boundary of inter-agent communication.

Some works occupy the boundary rather than the centre of this definition. Runtime systems such as TokenDance~\citep{bian2026tokendance} and AAFLOW+~\citep{sarker2026aaflow} reuse or orchestrate KV state across agent workflows but do not necessarily define a new semantic message protocol. Conversely, internalised-debate methods such as Latent Agents~\citep{yi2026latentagents} move multi-agent structure into a single model at deployment time. We discuss these as adjacent evidence rather than silently merging them with direct sender--receiver protocols.

\subsection{Extraction and Coding}

For each primary work, we record the transmitted representation, sender--receiver compatibility assumptions, layer mapping, fusion operator, training requirement, communication topology, evaluation tasks, and reported efficiency/quality metrics. We then code each method along the WHAT / WHICH / HOW axes. When a paper spans multiple choices, the comparison table reports the dominant inference path and the method card records variants. Reported numerical results are retained as \emph{within-paper} evidence; we do not treat them as a leaderboard because backbones, hardware, context lengths, baselines, and latency definitions differ.

\begin{takeaway}[Scope Principle]
This survey distinguishes three objects that are often conflated: \textbf{semantic latent messages} exchanged to improve collaboration, \textbf{reusable inference state} exchanged to avoid computation, and \textbf{internalised multi-agent reasoning} distilled into one model. They share mechanisms but answer different research questions.
\end{takeaway}

\section{Background and Preliminaries}
\label{sec:prelim}

This section fixes the notation and terminology used throughout the paper.

\subsection{Multi-Agent LLM Systems}

A \emph{multi-agent LLM system} (MAS) consists of $N$ LLM agents $\mathcal{A} = \{A_1, A_2, \ldots, A_N\}$, each equipped with a role-specific system prompt, optional tool access, and a communication channel. At each step, an agent $A_i$ (the \emph{sender}) produces a message that is delivered to one or more peer agents (the \emph{receivers}). A controller --- explicit or implicit --- decides the order of speakers. The communication channel is the focus of this paper: classical systems use a \emph{natural language channel} (\Cref{sec:prelim:nl}); the methods surveyed in this paper use a \emph{latent channel} (\Cref{sec:prelim:latent}).

\subsection{Natural Language vs.\ Latent Communication}
\label{sec:prelim:nl}
\label{sec:prelim:latent}

\begin{itemize}[leftmargin=*, itemsep=2pt]
  \item \textbf{Natural Language Communication (NL-Comm).} The sender generates a discrete token sequence $y = (y_1, y_2, \ldots, y_T)$ by sampling from a vocabulary $\mathcal{V}$. The receiver \emph{re-encodes} the token sequence into its own embedding space. The two-step pipeline --- \emph{sender decode $\rightarrow$ token transport $\rightarrow$ receiver encode} --- is what we refer to as the \emph{language bottleneck}.
  \item \textbf{Latent Communication (Latent-Comm).} The sender exposes one of its internal continuous representations --- the input embedding, the hidden state of a particular layer/token, or the KV-cache --- and the receiver injects this representation into its own computation \emph{without} round-tripping through the vocabulary.
\end{itemize}

\begin{figure}[t]
  \centering
  \includegraphics[width=0.95\linewidth]{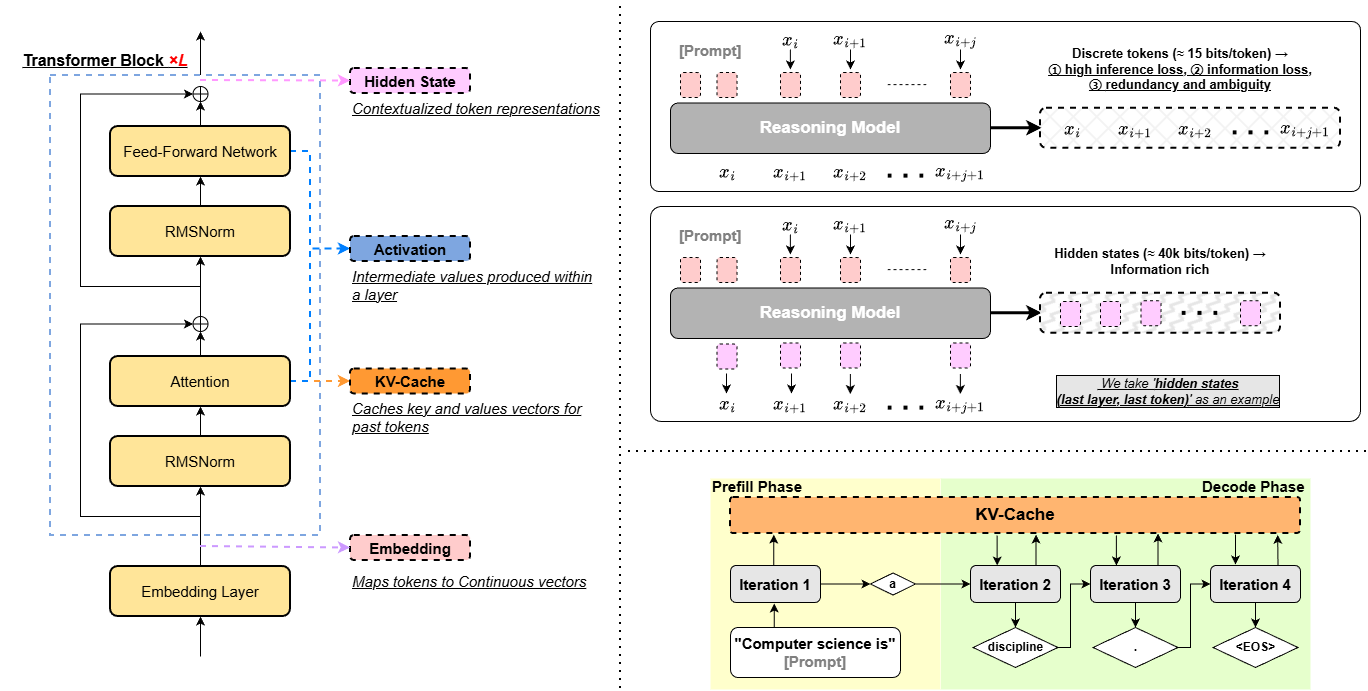}
  \caption{Comparison of natural-language and latent communication pipelines, including (\emph{left}) a Transformer block with its accessible intermediate representations, (\emph{top-right}) a comparison of token-level vs.\ hidden-state reasoning information density, and (\emph{bottom-right}) the prefill/decode phases that produce per-token KV-caches.}
  \label{fig:preliminary}
\end{figure}

A high-level comparison of the two pipelines is shown in \Cref{fig:preliminary}.

\subsection{Prefill and Decode}

LLM inference is split into two phases that we will repeatedly refer to:

\begin{itemize}[leftmargin=*, itemsep=2pt]
  \item \textbf{Prefill phase.} Given a prompt $x = (x_1, \ldots, x_T)$, the model processes the entire sequence in parallel and produces the first output token. All key--value pairs computed during prefill are stored in the KV-cache.
  \item \textbf{Decode phase.} The model generates one token at a time. At each step $t > T$, it takes the previously generated token $y_{t-1}$ and the cached KV from earlier steps, and produces a new token $y_t$ (and a new KV entry).
\end{itemize}

The distinction matters for latent communication because the \emph{kind} of internal state available differs between the two phases. During prefill, the sender has access to per-token hidden states and KV-caches for \emph{every} input token. During decode, the sender has only the hidden state of the most recently generated token plus an ever-growing KV-cache.

\subsection{Embedding, Hidden State, KV-Cache, Activation}

We adopt the following precise definitions, which the rest of the paper relies on:

\begin{description}[leftmargin=1.5em, style=nextline, itemsep=2pt]
  \item[\textbf{Embedding.}] A continuous vector $\mathbf{e}_i \in \mathbb{R}^d$ that maps a discrete input symbol $x_i$ to a dense semantic space. Embeddings are the \emph{input} to the first Transformer block.
  \item[\textbf{Hidden state.}] The output of a complete Transformer block, denoted $\mathbf{h}_i^{(\ell)} \in \mathbb{R}^d$ for token $i$ at layer $\ell$. Hidden states are the \emph{stable, layer-wise semantic representations} passed between adjacent Transformer blocks. When the receiver consumes a hidden state, it typically receives one of the intermediate-layer outputs.
  \item[\textbf{KV-Cache.}] The collection of per-token key and value tensors computed in each self-attention layer during prefill, denoted $\mathcal{KV} = \{ (\mathbf{k}_i^{(\ell)}, \mathbf{v}_i^{(\ell)})_{i=1}^{T} \}_{\ell=1}^{L}$. The KV-cache is what the model reuses to make decode efficient.
  \item[\textbf{Activation.}] A more general term: any intermediate output of a sub-module (attention projection, feed-forward transformation, etc.). \emph{Hidden states are a subset of activations that serve as stable layer-wise representations}. Methods that transmit ``activations'' often transmit a more granular quantity (e.g., a single attention head's output) than methods that transmit ``hidden states.''
\end{description}

A schematic of these quantities in a Transformer block is included in the left panel of \Cref{fig:preliminary}.

\subsection{Why Now?}

Latent communication has become practical only recently. Three enabling trends converged around 2023--2024:

\begin{enumerate}[leftmargin=*, itemsep=2pt]
  \item \textbf{Open-weight LLMs at scale.} Methods that pipe a sender's hidden state into a receiver's forward pass require \emph{white-box} access to both models. The release of Llama, Qwen, Mistral, and similar families has made such access routine.
  \item \textbf{KV-cache engineering.} The KV-cache has gone from an implementation detail to a first-class optimisation target, with rich infrastructure for compression, sharing, and off-loading. Methods that transmit KV-caches piggy-back on this infrastructure.
  \item \textbf{Multi-agent frameworks.} Frameworks like LangGraph, AutoGen, CrewAI, and MetaGPT have lowered the cost of orchestrating multiple LLM agents, making the \emph{latent channel} itself a meaningful object of study rather than a curiosity.
\end{enumerate}

\section{The Case for Latent Communication}
\label{sec:case}

Before diving into the framework, we articulate the case \emph{for} and \emph{against} latent communication. We argue that the trade-off is context-dependent: latent communication is preferable when (a) the agents are tightly coupled, (b) the cost of natural language overhead dominates, and (c) the channel can be made interpretable enough for downstream debugging.

\subsection{Limitations of Natural Language Communication}

\subsubsection{High Inference Cost}

Consider a two-agent system where agent $A_1$ produces a $T$-token message to agent $A_2$. The total cost is:
\begin{itemize}[leftmargin=*, itemsep=2pt]
  \item $A_1$'s decode of $T$ tokens: $\mathcal{O}(L \cdot T \cdot d)$ FLOPs, where $L$ is the number of layers and $d$ is the hidden dimension. The KV-cache read/write is the dominant term.
  \item $A_2$'s re-encoding of $T$ tokens: the same $\mathcal{O}(L \cdot T \cdot d)$ FLOPs in prefill.
  \item The token-by-token transport itself: negligible.
\end{itemize}
So the \emph{overhead} of natural language communication is roughly $2 \times$ the cost of generating the message, even before accounting for $A_2$'s own reasoning. Latent communication can reduce this to a single embedding/hidden-state/KV-cache transport that the receiver injects \emph{without re-encoding}.

\subsubsection{Information Loss During Discretization}

The pipeline is
\begin{equation}
  \mathbf{h}_{\text{context}} \xrightarrow{\text{linear}} \mathbf{z} \in \mathbb{R}^{V} \xrightarrow{\text{sample}} y \in \mathcal{V},
  \label{eq:discretisation}
\end{equation}
where $\mathbf{h}_{\text{context}}$ is the sender's high-dimensional hidden state, $\mathbf{z}$ is the logit vector over the vocabulary, and $y$ is the sampled token. The mutual information $I(\mathbf{h}_{\text{context}}; y)$ is upper-bounded by $H(y) \le \log_2 |\mathcal{V}| \approx 15\text{--}17$ bits. Meanwhile, $\mathbf{h}_{\text{context}}$ itself typically lives in $\mathbb{R}^d$ with $d \ge 4096$ and is parameterised by 32-bit floats, so its \emph{raw} representational capacity exceeds $40{,}000$ bits. The compression factor is therefore on the order of $10^{3}$--$10^{4}$.

\begin{figure}[t]
  \centering
  \begin{subfigure}[t]{0.48\linewidth}\centering
    \includegraphics[width=\linewidth]{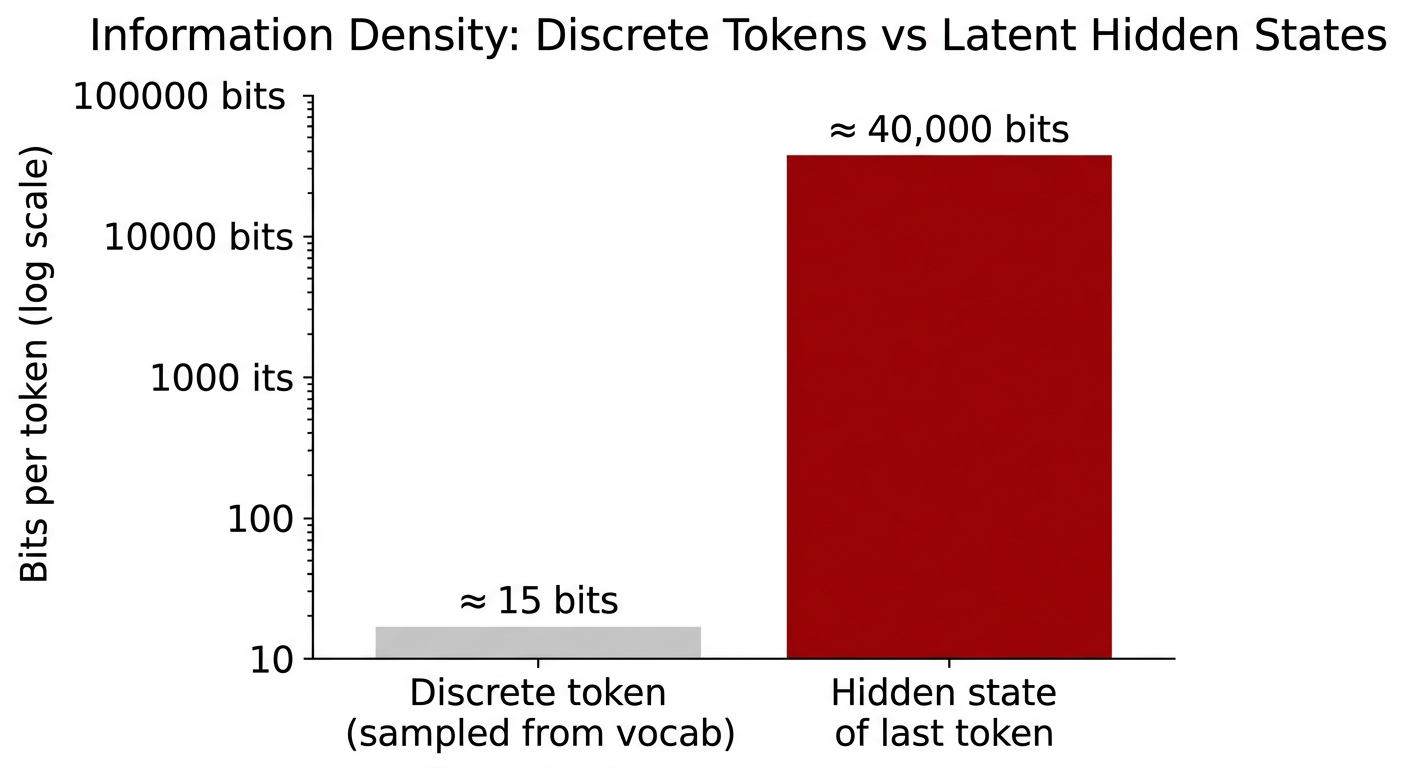}
    \caption{Information density: $\approx 15$ bits per token.}\label{fig:infodensity-a}
  \end{subfigure}\hfill
  \begin{subfigure}[t]{0.48\linewidth}\centering
    \includegraphics[width=\linewidth]{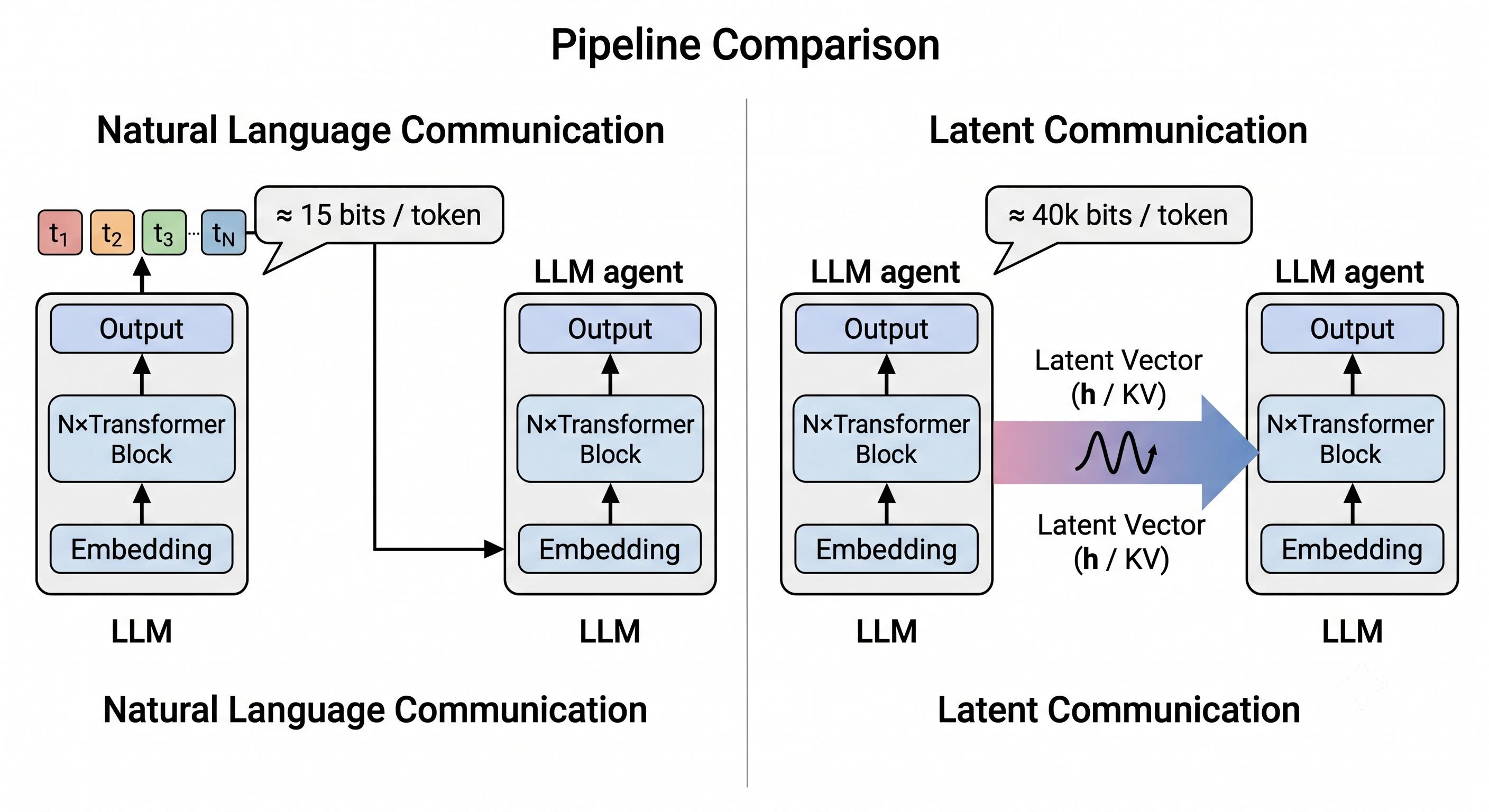}
    \caption{Pipeline comparison: NL-Comm vs.\ Latent-Comm.}\label{fig:infodensity-b}
  \end{subfigure}
  \caption{Why latent communication wins on information density. \emph{Left (a):} Bar chart comparing the information content of a discrete token ($\approx 15$ bits) with that of a single hidden state of the last token ($\approx 40{,}000$ bits). The gap of three to four orders of magnitude motivates the move to latent communication. \emph{Right (b):} Pipeline comparison. NL-Comm routes a sender's hidden state through a vocabulary bottleneck; Latent-Comm exchanges a continuous vector directly, preserving orders of magnitude more information per communication step.}
  \label{fig:infodensity}
\end{figure}

Concretely: a hidden state encodes not just \emph{which} token to say next, but also the \emph{alternatives} considered, their \emph{relative probabilities}, the \emph{salience} of different parts of the context, and \emph{uncertainty}. All of this is lost the moment we sample a single token. A visual comparison of these information densities (\Cref{fig:infodensity}(a)) and the resulting communication pipelines (\Cref{fig:infodensity}(b)) is given in \Cref{fig:infodensity}.

\subsubsection{Redundancy and Ambiguity of Natural Language}

Generated text is optimised for linguistic coherence (a stylistic objective from pre-training) rather than for \emph{task-relevant information density}. Sentences are padded with politeness markers, hedging, and reformulation. References to prior context are often under-specified (``the previous step'', ``that approach''), forcing the receiver to reconstruct the referent.

When sender and receiver disagree on background knowledge, role assignment, or terminology, the natural language channel can become lossy in a \emph{semantic} sense that goes beyond the numerical bits/token argument. In contrast, latent channels operate on the agents' own representational manifolds and avoid this kind of semantic mismatch --- at the cost of interpretability.

\subsection{Advantages of Natural Language Communication}

Latent communication is not a universal replacement. Natural language retains one decisive advantage:

\begin{itemize}[leftmargin=*, itemsep=2pt]
  \item \textbf{High interpretability.} A natural language message is immediately readable by humans. This is essential for \emph{debugging}, \emph{alignment auditing}, \emph{safety review}, and \emph{human--AI interaction}. Latent messages, in contrast, are opaque vectors that require auxiliary tooling to interpret.
\end{itemize}

In practice, the field has converged on a hybrid view: natural language for tasks where human oversight is needed (e.g., final answers, justifications) and latent communication for intermediate, agent-to-agent signalling. A schematic of this hybrid view is shown in \Cref{fig:infodensity}(b).

\subsection{When to Prefer Latent Communication}

Synthesising the above, latent communication tends to win when \emph{all} of the following hold:
\begin{enumerate}[leftmargin=*, itemsep=2pt]
  \item The two agents are \emph{tightly coupled} (e.g., a planner feeding directly into an executor).
  \item The communication is \emph{intermediate} --- the user does not need to see the message.
  \item The sender and receiver share (or can be aligned to) a \emph{common latent space} (e.g., same backbone, or compatible architectures).
  \item \emph{Latency} is a binding constraint (e.g., real-time pipelines, edge deployment, or large agent counts).
\end{enumerate}
Conversely, natural language wins when interpretability, cross-organisation interoperability, or human oversight is required.

\section{A Unified Framework for Latent Communication}
\label{sec:framework}

We now present the central contribution of this paper: a \emph{unified framework} that describes the core data path of latent communication methods along three axes. A protocol can be represented by the triple
\begin{equation}
  \text{Method} = (\underbrace{\text{WHAT}}_{\text{type of information}},\ \underbrace{\text{WHICH}}_{\text{alignment}},\ \underbrace{\text{HOW}}_{\text{fusion}}).
  \label{eq:triple}
\end{equation}

\begin{figure}[t]
  \centering
  \includegraphics[width=0.8\linewidth]{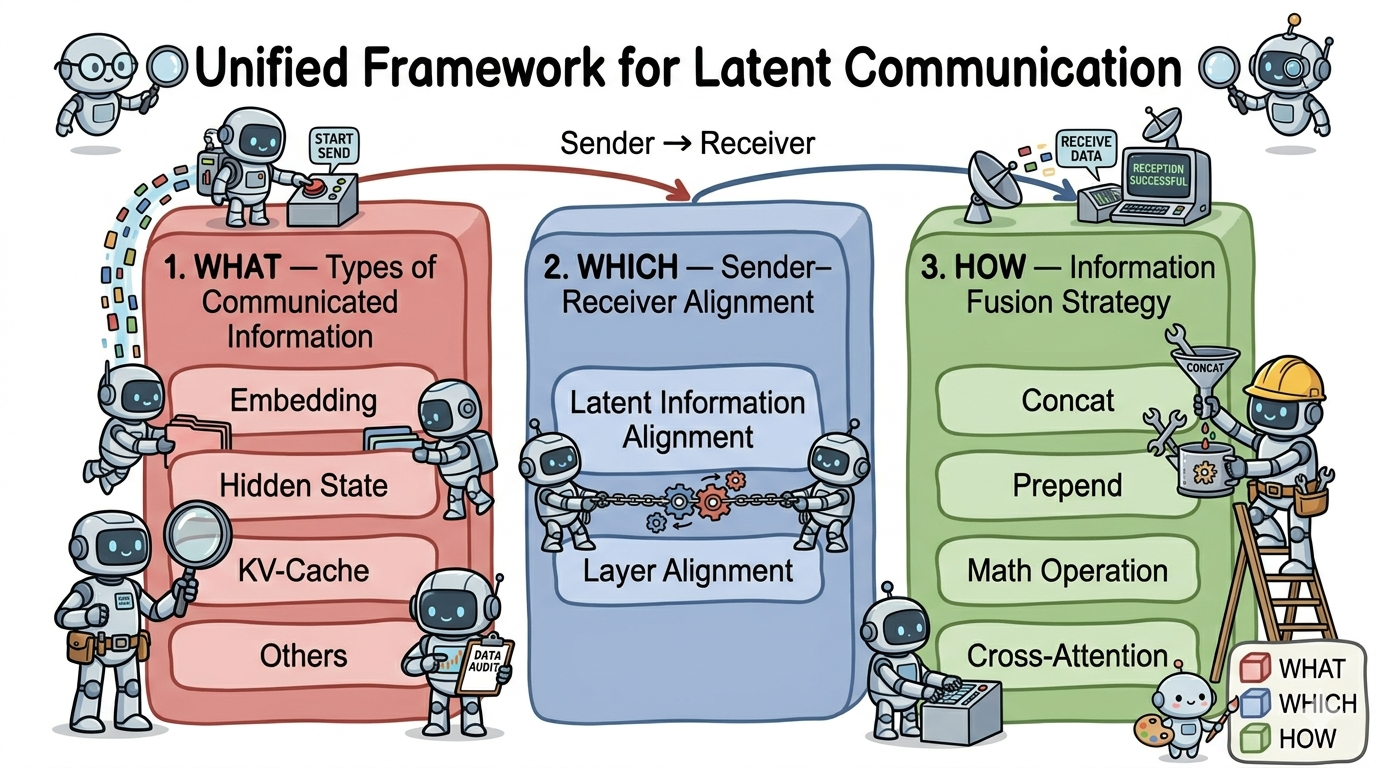}
  \caption{The unified 3-axis framework. The three axes --- \textbf{WHAT} (types of communicated information), \textbf{WHICH} (sender--receiver alignment), and \textbf{HOW} (information fusion strategy) --- together span the design space of latent communication methods.}
  \label{fig:framework-overview}
\end{figure}

The framework is summarised schematically in \Cref{fig:framework-overview}.

\begin{highlightbox}[At a glance]
  \textbf{WHERE} does the information come from? \textbf{WHAT} is its format? \textbf{WHICH} layer/head in the receiver does it target? \textbf{HOW} is it combined? These three questions (\textbf{WHAT / WHICH / HOW}) uniquely determine any latent communication protocol.
\end{highlightbox}

\subsection{Axis 1 --- \textbf{WHAT}: Types of Communicated Information}
\label{sec:what}

The first axis asks: \emph{what continuous quantity does the sender expose to the receiver?} The dominant choices in the literature are \textbf{Embeddings}, \textbf{Hidden States}, and \textbf{KV-Caches}, with several methods exploring \emph{other} quantities (state deltas, persistent memory, attention-only signals).

\subsubsection{Embeddings}

The sender transmits its input embedding $\mathbf{e}_i \in \mathbb{R}^d$ for one or more tokens. Embeddings are the lowest-level continuous representation; they are model-agnostic in the sense that \emph{any} model with a compatible embedding dimension can in principle consume them. \method{CIPHER}~\citep{liu2024cipher} is the canonical example: it computes a \emph{weighted} embedding where the weights are derived from the sender's output logits, so that the embedding encodes the sender's \emph{full} vocabulary distribution rather than a single sampled token.

\begin{description}[leftmargin=1.5em, style=nextline, itemsep=2pt]
  \item[Strengths.] Architecture-light (only the embedding table needs to be shared). Simple to implement. Robust to backbone changes.
  \item[Limitations.] Embeddings are the \emph{least informative} of the three options. They do not encode the agent's intermediate computations or its attended context.
\end{description}

\subsubsection{Hidden States}

The sender transmits the hidden state $\mathbf{h}_i^{(\ell)}$ of token $i$ at layer $\ell$. Hidden states are richer than embeddings: they encode the agent's \emph{intermediate} reasoning, including the effect of attention over its context. \method{AC}~\citep{ye2025ac}, \method{Interlat}~\citep{du2026interlat}, \method{SDE}~\citep{yang2025sde}, \method{ThoughtComm}~\citep{li2025thoughtcomm}, and \method{Mixture of Thoughts}~\citep{feinashley2025mot} all use hidden states as the communicated quantity.

\begin{description}[leftmargin=1.5em, style=nextline, itemsep=2pt]
  \item[Strengths.] Encodes intermediate computation. Often training-free. Easy to align to the receiver's first layer (\Cref{sec:which}).
  \item[Limitations.] Less informative than the full KV-cache (it does not include the keys needed to attend back to earlier tokens). Architecture-dependent: the receiver must share a similar backbone.
\end{description}

\subsubsection{KV-Caches}

The sender transmits its per-token, per-layer KV-cache. The receiver can then \emph{resume} generation as if it had pre-filled the sender's context. \method{KVComm}~\citep{wang2025kvcomm}, \method{Cache-to-Cache}~\citep{liu2025c2c}, \method{LatentMAS}~\citep{wang2025latentmas}, \method{Q-KVComm}~\citep{park2025qkvcomm}, \method{LRAgent}~\citep{jeon2026lra}, \method{RelayCaching}~\citep{geng2026relay}, \method{Agent Memory}~\citep{shkolnikov2026am}, \method{Agent Primitives}~\citep{jin2026primitives}, and \method{Edge LLM Handover}~\citep{lee2026edge} all use KV-caches.

\begin{description}[leftmargin=1.5em, style=nextline, itemsep=2pt]
  \item[Strengths.] Retains token- and layer-specific attention state, allowing the receiver to attend over a richer history than a single hidden vector. Compatible with existing KV-cache infrastructure.
  \item[Limitations.] Largest payload (proportional to sequence length $\times$ number of layers $\times$ number of heads $\times$ head dimension). Most architecture-dependent: a KV-cache from a 4096-d Llama cannot be directly consumed by a 5120-d Qwen. Requires careful alignment across architectures.
\end{description}

\subsubsection{Other Communicated Quantities}

A small but growing set of methods transmits \emph{non-standard} quantities:
\begin{itemize}[leftmargin=*, itemsep=2pt]
  \item \textbf{State delta trajectory} (\method{SDE}~\citep{yang2025sde}): the \emph{change} in hidden state at each layer, rather than the state itself. This compresses the information into a direction in latent space and has been shown to be more robust when sender and receiver architectures differ slightly.
  \item \textbf{Persistent KV-cache memory} (\method{Agent Memory}~\citep{shkolnikov2026am}): a disk-persistent 4-bit-quantised KV-cache, used to offload the cache to edge devices.
  \item \textbf{Visual-latent wormhole} (\method{Vision Wormhole}~\citep{liu2026vision}): a sender's hidden state is \emph{rendered} into a VLM's visual input space, exploiting the VLM's visual pathway as a universal channel.
  \item \textbf{Centralised workspace state} (\method{BIGMAS}~\citep{hao2026bigmas}): a shared workspace in which agents deposit and read structured latent messages, mediated by an orchestrator.
\end{itemize}

\subsubsection{Comparative Summary}

\Cref{tab:what} summarises which information type each primary method uses and places the four recently added boundary works in the same representation space.

\begin{table}[t]
  \centering
  \footnotesize
  \caption{Types of communicated information used by representative methods ($\checkmark$ = yes). $^\dagger$ denotes boundary or adjacent work not counted among the 18 primary methods.}
  \label{tab:what}
  \begin{tabular}{lcccc}
    \toprule
    \textbf{Method} & \textbf{Embedding} & \textbf{Hidden State} & \textbf{KV-Cache} & \textbf{Other} \\
    \midrule
    CIPHER~\citep{liu2024cipher}            & $\checkmark$ (weighted) &                  &                  &         \\
    AC~\citep{ye2025ac}                    &                         & $\checkmark$      &                  &         \\
    Interlat~\citep{du2026interlat}        &                         & $\checkmark$      &                  &         \\
    SDE~\citep{yang2025sde}                &                         &                  &                  & delta   \\
    KVComm~\citep{wang2025kvcomm}          &                         &                  & $\checkmark$      &         \\
    Cache-to-Cache~\citep{liu2025c2c}      &                         &                  & $\checkmark$      &         \\
    LatentMAS~\citep{wang2025latentmas}    &                         &                  & $\checkmark$      &         \\
    ThoughtComm~\citep{li2025thoughtcomm}  &                         & $\checkmark$      &                  &         \\
    Q-KVComm~\citep{park2025qkvcomm}       &                         &                  & $\checkmark$ (comp.) &    \\
    LRAgent~\citep{jeon2026lra}            &                         &                  & $\checkmark$ (base+lr) &  \\
    RelayCaching~\citep{geng2026relay}     &                         &                  & $\checkmark$ (dec)  &      \\
    Agent Memory~\citep{shkolnikov2026am}  &                         &                  & $\checkmark$ (Q4)   &      \\
    Agent Primitives~\citep{jin2026primitives} &                     &                  & $\checkmark$ (inter)  &      \\
    Edge LLM Handover~\citep{lee2026edge}  &                         &                  & $\checkmark$ (backhaul) &    \\
    Vision Wormhole~\citep{liu2026vision}  &                         &                  &                  & UVC     \\
    BIGMAS~\citep{hao2026bigmas}           &                         &                  &                  & workspace \\
    Mixture of Thoughts~\citep{feinashley2025mot} &                 & $\checkmark$ (proj.) &               &         \\
    Five Ws Survey~\citep{chen2026fivews}  & (all)                   & (all)            & (all)            & (taxonomy) \\
    \midrule
    TokenDance$^\dagger$~\citep{bian2026tokendance} &                  &                    & $\checkmark$ (shared/diff) & runtime reuse \\
    AAFLOW+$^\dagger$~\citep{sarker2026aaflow}      &                  &                    & $\checkmark$ (distributed) & operators \\
    When Latent Agents Lie$^\dagger$~\citep{brito2026latentlie} &      &                    & $\checkmark$ (full) & visible text \\
    Latent Agents$^\dagger$~\citep{yi2026latentagents} &               & $\checkmark$ (subspaces) &               & internalised \\
    \bottomrule
  \end{tabular}
\end{table}

\begin{takeaway}[Takeaway 1 --- Information--Cost--Dependence Trade-off]
Payload size and architecture dependence generally increase from embeddings to selected hidden states to full KV-caches. Useful task information need not follow the same strict ordering: alignment error, token selection, and fusion can make a smaller representation more effective than a larger one.
\end{takeaway}

\begin{takeaway}[Takeaway 2 --- Prefill vs.\ Decode Choice]
KV-Cache methods are concentrated in the \emph{prefill} phase (because the cache is naturally produced there), while Embedding/Hidden State methods are concentrated in the \emph{decode} phase (because the last-token hidden state is what the model uses to predict the next token). When the receiver only consumes prefill-phase information, the \emph{sender's decode phase can be skipped entirely}, yielding a major inference speed-up --- the key insight behind all KV-cache methods.
\end{takeaway}

\subsection{Axis 2 --- \textbf{WHICH}: Sender--Receiver Alignment}
\label{sec:which}

The second axis asks: \emph{which parts of the sender correspond to which parts of the receiver?} Alignment has two sub-dimensions: \emph{latent information alignment} (does the sender's semantic space match the receiver's?) and \emph{layer alignment} (which layer of the sender feeds into which layer of the receiver?).

\subsubsection{Latent Information Alignment}

If the sender and receiver are \emph{the same model} (e.g., two instances of Llama-3-8B), their latent spaces are \emph{identical by construction} --- no alignment is needed. If they are \emph{different} but architecturally compatible (e.g., two Llama-3 fine-tunes), the spaces are \emph{close} but not identical; methods such as \method{Interlat}~\citep{du2026interlat} and \method{Cache-to-Cache}~\citep{liu2025c2c} apply \emph{learned} projection heads to bridge the gap. If they are \emph{architecturally heterogeneous} (e.g., Llama-3 and Qwen-2), a \emph{Universal Visual Codec}~\citep{liu2026vision} or a \emph{learned interaction layer}~\citep{feinashley2025mot} is needed.

\Cref{tab:which} indicates which methods perform explicit alignment.

\begin{table}[t]
  \centering
  \footnotesize
  \setlength{\tabcolsep}{4pt}
  \caption{Sender--receiver alignment assumptions. $^\dagger$ denotes boundary or adjacent work.}
  \label{tab:which}
  \begin{tabular}{@{}p{0.28\linewidth}p{0.60\linewidth}@{}}
    \toprule
    \textbf{Method} & \textbf{Latent Information Alignment} \\
    \midrule
    CIPHER, AC, SDE, KVComm, Q-KVComm, LRAgent, RelayCaching, Agent Memory, Agent Primitives, Edge LLM Handover, ThoughtComm, BIGMAS & None (same model assumed) \\
    Interlat, Cache-to-Cache, LatentMAS & Learned projection (homogeneous backbone) \\
    Vision Wormhole & Universal Visual Codec (heterogeneous) \\
    Mixture of Thoughts & Interaction layers (heterogeneous) \\
    TokenDance$^\dagger$, AAFLOW+$^\dagger$ & Same-checkpoint cache compatibility plus position/runtime metadata \\
    When Latent Agents Lie$^\dagger$ & Same-model KV payload; integrity binding rather than semantic alignment \\
    Latent Agents$^\dagger$ & Intra-model agent subspaces; no inter-agent alignment at deployment \\
    \bottomrule
  \end{tabular}
\end{table}

\subsubsection{Layer Alignment}

The second sub-axis specifies the \emph{layer-to-layer} correspondence between sender and receiver. Two natural extremes appear repeatedly:

\begin{itemize}[leftmargin=*, itemsep=2pt]
  \item \textbf{Last $\rightarrow$ First.} The sender exposes the hidden state of its \emph{last} layer, and the receiver injects it at its \emph{first} layer. Used by CIPHER, AC, Interlat. This is the simplest mapping and works well when the sender's last layer is the most semantically rich.
  \item \textbf{All $\rightarrow$ Corresponding.} The sender exposes the hidden state of \emph{every} layer, and the receiver injects each one into the \emph{corresponding} layer (i.e., layer $\ell$ of the sender feeds layer $\ell$ of the receiver). Used by Cache-to-Cache, LatentMAS, SDE. This preserves the layer-wise structure of the sender's computation and is the natural choice for KV-cache methods.
\end{itemize}

Intermediate variants include:
\begin{itemize}[leftmargin=*, itemsep=2pt]
  \item \textbf{Selected $\rightarrow$ Selected.} The sender selects $n \ge 1$ layers via a heuristic or learned gate, and the receiver injects them at the same indices. Used by AC, KVComm, Q-KVComm.
  \item \textbf{Sparse top-$k$ attention.} A sub-variant of selected $\rightarrow$ selected in which the receiver attends over only the top-$k$ most relevant layers (used by KVComm).
\end{itemize}

\begin{takeaway}[Takeaway 3 --- Layer Mapping Strategies]
For \emph{homogeneous} agents (same backbone), both the ``last $\rightarrow$ first'' and ``all $\rightarrow$ corresponding'' strategies are simple, training-free, and competitive. The ``selected $\rightarrow$ selected'' strategy adds complexity but can yield accuracy or latency gains when the agent has many layers and the relevant information is concentrated in a few. For \emph{heterogeneous} agents, learned alignment (projection, universal codec, or interaction layers) becomes necessary.
\end{takeaway}

\subsection{Axis 3 --- \textbf{HOW}: Information Fusion Strategy}
\label{sec:how}

The third axis asks: \emph{how is the communicated information incorporated into the receiver's computation?} The major options are:

\subsubsection{Concatenation}

The sender's latent is concatenated with the receiver's prompt embedding (or hidden state) along the token axis. This is the simplest fusion and is used by CIPHER, Interlat, and several early hidden-state methods.

\subsubsection{Prepending (Token-axis Prepend)}

The sender's latent is \emph{prepended} to the receiver's KV-cache. This is the natural fusion for KV-cache methods: the receiver can attend over the sender's context as if it were the first few tokens of its own prompt. Used by KVComm, LatentMAS, and others.

\subsubsection{Mathematical Operation}

The sender's latent is \emph{combined} with the receiver's hidden state (or KV-cache) by an element-wise operation: addition, subtraction, or a small learned linear projection. Used by AC (addition of last-token hidden states), SDE (addition of state deltas), and others.

\subsubsection{Cross-Attention}

The receiver attends over a \emph{set} of sender latents using a learned cross-attention layer. Used by \method{Mixture of Thoughts}~\citep{feinashley2025mot}, where a primary expert attends over a top-$K$ set of peer experts' projected hidden states.

\subsubsection{Cache Restoration / Direct Injection}

The receiver \emph{replaces} part of its own KV-cache with the sender's KV-cache. Used by RelayCaching~\citep{geng2026relay} and Agent Memory~\citep{shkolnikov2026am}, where the goal is to \emph{avoid} recomputation rather than to mix information.

\subsubsection{Comparative Table}

\Cref{tab:how} lists the fusion strategy for the primary methods and the newly added boundary works.

\begin{table}[t]
  \centering
  \footnotesize
  \caption{Fusion strategies by method. $^\dagger$ denotes a boundary or adjacent work not counted among the 18 primary methods.}
  \label{tab:how}
  \begin{tabular}{p{0.32\linewidth}p{0.32\linewidth}p{0.27\linewidth}}
    \toprule
    \textbf{Method} & \textbf{Fusion Strategy} & \textbf{Detail} \\
    \midrule
    CIPHER~\citep{liu2024cipher}            & Concatenation  & Weighted embedding concat.\ at each step \\
    AC~\citep{ye2025ac}                    & Math operation & Hidden state of last token added/combined \\
    Interlat~\citep{du2026interlat}        & Concatenation  & Latent appended to prompt embeddings \\
    SDE~\citep{yang2025sde}                & Math operation & State delta added to corresponding layer/token \\
    KVComm~\citep{wang2025kvcomm}          & Prepend        & Sender KV prepended to receiver KV \\
    Cache-to-Cache~\citep{liu2025c2c}      & Math operation & Learned fuser \\
    LatentMAS~\citep{wang2025latentmas}    & Prepend        & Sender KV prepended to receiver KV \\
    ThoughtComm~\citep{li2025thoughtcomm}  & Math operation & Hidden-state combination \\
    Q-KVComm~\citep{park2025qkvcomm}       & Prepend        & Compressed KV prepended \\
    LRAgent~\citep{jeon2026lra}            & Math operation & Base + low-rank adapter inside fused kernel \\
    RelayCaching~\citep{geng2026relay}     & Cache restoration & Direct transplant + sparse recomputation \\
    Agent Memory~\citep{shkolnikov2026am}  & Cache restoration & Q4 disk-resident cache reloaded into attention \\
    Agent Primitives~\citep{jin2026primitives} & Prepend    & Inter-primitive KV exchange \\
    Edge LLM Handover~\citep{lee2026edge}  & Hybrid         & Partial re-prefill + partial KV transfer \\
    Vision Wormhole~\citep{liu2026vision}  & Visual injection & Latent rendered into VLM's visual input space \\
    BIGMAS~\citep{hao2026bigmas}           & Workspace fusion & Orchestrator routes full shared state \\
    Mixture of Thoughts~\citep{feinashley2025mot} & Cross-attention & Primary expert attends over top-$K$ peer hidden states \\
    \midrule
    TokenDance$^\dagger$~\citep{bian2026tokendance} & Collective cache reuse & Master--Mirror sparse-difference restoration \\
    AAFLOW+$^\dagger$~\citep{sarker2026aaflow} & Cache orchestration & Materialise, transfer, fork, compose, evict, and restore \\
    When Latent Agents Lie$^\dagger$~\citep{brito2026latentlie} & KV injection + integrity & Full-KV coordination with a cryptographically bound manifest \\
    Latent Agents$^\dagger$~\citep{yi2026latentagents} & Post-training internalisation & No inter-agent receiver fusion at deployment \\
    \bottomrule
  \end{tabular}
\end{table}

\begin{takeaway}[Takeaway 4 --- Fusion Strategy Spectrum]
\emph{Concatenation} and \emph{prepending} are the most common, simplest, and often training-free. \emph{Mathematical operations} (addition, learned linear) are slightly more expressive but require architectural compatibility. \emph{Cross-attention} is the most expressive but requires training. \emph{Cache restoration} is the most efficient (avoids recomputation entirely) but is the most restrictive in scope.
\end{takeaway}

\subsection{Combining the Axes}

The three axes are conceptually separable, although not fully independent in implementations: KV-caches favour layer-wise injection, whereas embeddings naturally enter near the input. Their combinations nevertheless reveal unexplored regions, such as compressed heterogeneous caches and auditable multi-sender latent aggregation.

\subsection{From Taxonomy to Design Decisions}
\label{sec:design-guidance}

The axes describe a protocol, but system builders need the inverse mapping: given a workload, which point in the design space is defensible? \Cref{tab:decision-guide} summarises the strongest recurring correspondences. These are design heuristics rather than universal rankings; they should be validated under the target model, hardware, and communication topology.

\begin{table}[t]
  \centering
  \footnotesize
  \setlength{\tabcolsep}{4pt}
  \caption{Decision guide linking deployment constraints to protocol choices.}
  \label{tab:decision-guide}
  \begin{tabular}{@{}p{0.25\linewidth}p{0.25\linewidth}p{0.40\linewidth}@{}}
    \toprule
    \textbf{Dominant constraint} & \textbf{Promising choice} & \textbf{Rationale and caveat} \\
    \midrule
    Same backbone; long shared context & KV-cache transfer or restoration & Avoids repeated prefill, but payload and positional compatibility must be controlled. \\
    Heterogeneous backbones & Projected hidden states or shared codec & Smaller interface than full-cache translation, but alignment normally requires data or training. \\
    Tight bandwidth or edge deployment & Selected layers, quantisation, low-rank deltas & Reduces bytes transferred; quality must be measured after compression rather than inferred from reconstruction error alone. \\
    Strict auditability or external trust boundary & Hybrid text--latent channel & A concise visible commitment supports oversight, while the latent payload carries detail; the two must be cryptographically and semantically bound. \\
    Many senders or dense topology & Routing, pooling, or collective reuse & Prevents communication cost from scaling naively with every sender--receiver pair, but may suppress minority evidence. \\
    No training data or rapidly changing models & Same-model, training-free injection & Minimises adaptation cost, but does not remove distribution shift or implementation sensitivity. \\
    \bottomrule
  \end{tabular}
\end{table}

Three consequences follow. First, \emph{representation richness is not the same as useful information}: a larger KV payload may contain more state but can still harm the receiver if positions, layers, or model manifolds are misaligned. Second, the optimal unit of communication depends on topology. Pairwise hidden-state exchange may be adequate for a pipeline, whereas a dense debate may require routing or a shared workspace. Third, auditability is a protocol property, not an optional visualisation layer. Once agents communicate across trust boundaries, a pure latent channel should be treated as an untrusted binary interface with provenance, integrity, and fallback requirements.

\begin{takeaway}[Takeaway 5 --- Conditional, Not Universal, Superiority]
Latent communication is most compelling when it removes a measurable decode--encode or re-prefill bottleneck under compatible models. Natural language remains preferable when interoperability, human review, or organisational boundaries dominate. Hybrid protocols are therefore a first-class design point rather than a temporary compromise.
\end{takeaway}

\newcommand{\methodanalysisappendix}{%
\section{Method Analysis under the Framework}
\label{sec:methods}

This section provides a one-paragraph analysis for each of the eighteen methods, structured as: (a) core idea, (b) framework placement (WHAT / WHICH / HOW), (c) strengths, (d) limitations, and (e) reported results and code. Methods are grouped by the WHAT axis for narrative flow.

\subsection{Embedding-Based Methods}

\begin{methodbox}[CIPHER~\citep{liu2024cipher}]
\textbf{Core idea.} \emph{CIPHER} is an early method for communicating \emph{embeddings} rather than sampled tokens. The sender's output \emph{logits} over the vocabulary are converted to weights, and a weighted sum of embedding-table entries yields a \emph{soft embedding}. The receiver concatenates this representation to its prompt at each decode step.\\
\textbf{Framework.} WHAT = weighted Embedding. WHICH = last layer of sender $\rightarrow$ first layer of receiver. HOW = Concatenation.\\
\textbf{Strengths.} Training-free. Backbone-light (only the embedding table must be shared, not the rest of the model). Robust to model mismatches.\\
\textbf{Limitations.} Embeddings carry the \emph{least} information of the three options. Performance gains are modest compared to KV-cache methods.\\
\textbf{Results \& code.} Improves over token-level multi-agent debate on several reasoning and QA benchmarks. ICLR 2024. \href{https://github.com/chaudatascience/cipher_multiagent_debate}{Code}.
\end{methodbox}

\subsection{Hidden-State-Based Methods}

\begin{methodbox}[AC --- Communicating Activations~\citep{ye2025ac}]
\textbf{Core idea.} \method{AC} has the sender transmit the hidden state of the \emph{last token} at a \emph{selected layer} (typically a middle layer such as 16 in a 32-layer model). The receiver combines this hidden state with its own last-token hidden state via a simple mathematical operation (e.g., addition).\\
\textbf{Framework.} WHAT = Hidden State (last token, selected layer). WHICH = same layer in sender and receiver. HOW = Mathematical operation.\\
\textbf{Strengths.} Training-free. Encodes intermediate reasoning. Approximately 27\% accuracy improvement over natural language communication on a representative benchmark suite.\\
\textbf{Limitations.} Hidden state is not as informative as KV-cache. The choice of \emph{which} layer to transmit requires heuristic tuning.\\
\textbf{Results \& code.} Up to +27\% on math/reasoning benchmarks over NL-Comm. ICML 2025.
\end{methodbox}

\begin{methodbox}[Interlat~\citep{du2026interlat}]
\textbf{Core idea.} \method{Interlat} transmits the \emph{last-layer, last-token} hidden state from sender to receiver and concatenates it with the receiver's prompt embeddings. The receiver then proceeds with normal prefill. The authors introduce a small learned projection to align the sender's and receiver's last-layer spaces when the agents are \emph{different} fine-tunes of the same backbone.\\
\textbf{Framework.} WHAT = Hidden State. WHICH = last layer of sender $\rightarrow$ first layer of receiver, with optional learned projection. HOW = Concatenation.\\
\textbf{Strengths.} Simple, training-free for same-model agents. Up to $24\times$ speedup over NL-Comm on long-context multi-agent tasks.\\
\textbf{Limitations.} Last-layer / last-token state is less informative than the full cache. Performance depends on alignment quality.\\
\textbf{Results \& code.} Up to $24\times$ latency reduction on long-context tasks; competitive accuracy with NL-Comm. ACL 2026. \href{https://github.com/XiaoDu-flying/Interlat}{Code}.
\end{methodbox}

\begin{methodbox}[SDE --- State Delta Trajectory~\citep{yang2025sde}]
\textbf{Core idea.} \method{SDE} transmits not the hidden state itself, but the \emph{change} in hidden state at each layer (the ``state delta'') during a reasoning step. The deltas form a trajectory in latent space that, in the receiver, is \emph{added} to the corresponding layer/token of the receiver's hidden state.\\
\textbf{Framework.} WHAT = State-delta trajectory (a non-standard continuous quantity). WHICH = all layers. HOW = Mathematical operation (addition).\\
\textbf{Strengths.} Designed to be more robust to small architectural mismatches than raw hidden states; reports strong results on several complex reasoning benchmarks.\\
\textbf{Limitations.} The state-delta representation is unconventional and has not been widely adopted.\\
\textbf{Results \& code.} The paper reports improvements over its selected NL-Comm baselines. \href{https://github.com/LittleDinoC/StateDelta}{Code}.
\end{methodbox}

\begin{methodbox}[ThoughtComm~\citep{li2025thoughtcomm}]
\textbf{Core idea.} \method{ThoughtComm} treats each agent's \emph{thought} (intermediate hidden state) as a first-class message. The sender exposes its current hidden state; the receiver combines it with its own current hidden state through a learnable gating mechanism.\\
\textbf{Framework.} WHAT = Hidden State. WHICH = corresponding layer. HOW = Math operation (gated combination).\\
\textbf{Strengths.} Naturally aligned with the agent's \emph{internal} monologue. Works across homogeneous backbones.\\
\textbf{Limitations.} No public code at the time of writing.
\end{methodbox}

\begin{methodbox}[Mixture of Thoughts (MoT)~\citep{feinashley2025mot}]
\textbf{Core idea.} \method{MoT} is a heterogeneous latent communication method. A router selects a top-$K$ set of frozen LLM experts per query, and a primary expert performs cross-attention over projected peer hidden states. Uniformly placed interaction layers map each expert to a shared space, avoiding a separate translator for every ordered model pair.\\
\textbf{Framework.} WHAT = Hidden State (projected). WHICH = learned interaction layers. HOW = Cross-attention.\\
\textbf{Strengths.} Supports \emph{heterogeneous} experts and single-pass aggregation. The paper reports improvements over its comparison method on both in-distribution and out-of-distribution benchmarks.\\
\textbf{Limitations.} Requires training the router and interaction layers.\\
\textbf{Results \& code.} The paper reports average gains of 0.38\% on five in-distribution benchmarks and 2.92\% on three out-of-distribution benchmarks over Avengers. \href{https://github.com/jacobfa/mot}{Code}.
\end{methodbox}

\subsection{KV-Cache-Based Methods}

\begin{methodbox}[KVComm~\citep{wang2025kvcomm}]
\textbf{Core idea.} \method{KVComm} transmits a \emph{selected subset} of the sender's KV-cache (a few selected layers) to the receiver. Within the same layer index, the sender's KV is \emph{prepended} to the receiver's KV. A Gaussian-prior-based selection mechanism picks the most informative layers.\\
\textbf{Framework.} WHAT = KV-Cache (selected layers). WHICH = selected $\rightarrow$ corresponding. HOW = Prepend.\\
\textbf{Strengths.} Training-free. Reduces transmission cost relative to full-cache methods. Achieves strong latency improvements.\\
\textbf{Limitations.} Performance depends on the layer-selection heuristic.\\
\textbf{Results \& code.} Strong latency reduction on multi-agent QA pipelines. ICLR 2026.
\end{methodbox}

\begin{methodbox}[Cache-to-Cache (C2C)~\citep{liu2025c2c}]
\textbf{Core idea.} \method{Cache-to-Cache} transmits the \emph{entire} KV-cache from sender to receiver. All sender layers feed the corresponding receiver layers, and a small learned fuser blends the two caches.\\
\textbf{Framework.} WHAT = KV-Cache (all layers). WHICH = all $\rightarrow$ corresponding. HOW = Mathematical operation (learned fuser).\\
\textbf{Strengths.} Maximally informative. Strong empirical results on multi-agent reasoning.\\
\textbf{Limitations.} Highest transport cost among the surveyed methods. Fuser must be trained.\\
\textbf{Results \& code.} Significant accuracy gains on multi-agent benchmarks. \href{https://github.com/thu-nics/C2C}{Code}.
\end{methodbox}

\begin{methodbox}[LatentMAS~\citep{wang2025latentmas}]
\textbf{Core idea.} \method{LatentMAS} extends Cache-to-Cache by \emph{interleaving} prefill and decode: the sender's KV-cache is exposed \emph{both} during its prefill phase \emph{and} accumulated during its decode phase. The receiver gets the full prefill+decode KV-cache, prepended at every layer.\\
\textbf{Framework.} WHAT = KV-Cache (prefill + decode). WHICH = all $\rightarrow$ corresponding. HOW = Prepend.\\
\textbf{Strengths.} Maximally informative. Training-free. Strong results on collaborative reasoning.\\
\textbf{Limitations.} Largest transport cost.\\
\textbf{Results \& code.} The paper reports strong results on its collaborative reasoning benchmarks. \href{https://github.com/Gen-Verse/LatentMAS}{Code}.
\end{methodbox}

\begin{methodbox}[Q-KVComm~\citep{park2025qkvcomm}]
\textbf{Core idea.} \method{Q-KVComm} compresses the sender's KV-cache using an \emph{adaptive} quantisation scheme that achieves 5--6$\times$ compression while preserving semantic fidelity. The compressed cache is then transmitted to the receiver and prepended at the corresponding layers.\\
\textbf{Framework.} WHAT = KV-Cache (compressed). WHICH = all $\rightarrow$ corresponding. HOW = Prepend.\\
\textbf{Strengths.} Lowers transport cost by 5--6$\times$. Compatible with existing KV-cache infrastructure.\\
\textbf{Limitations.} Quantisation introduces small semantic drift.\\
\textbf{Results \& code.} 5--6$\times$ compression with negligible accuracy loss.
\end{methodbox}

\begin{methodbox}[LRAgent --- Multi-LoRA KV Sharing~\citep{jeon2026lra}]
\textbf{Core idea.} \method{LRAgent} addresses the \emph{multi-LoRA} setting: when several agents share the same backbone but use different LoRA adapters, the \emph{base} component of the KV-cache is identical across agents, while the \emph{adapter} component is small and low-rank. LRAgent shares the base component and stores the adapter component in low-rank form. A custom \emph{Flash-LoRA-Attention} kernel reconstructs adapter contributions without materialising the full cache.\\
\textbf{Framework.} WHAT = KV-Cache (base + low-rank adapter). WHICH = same backbone. HOW = Additive fusion inside fused kernel.\\
\textbf{Strengths.} Drastically reduces memory for multi-LoRA agents. Training-free at inference. Approximates fully-shared caching throughput.\\
\textbf{Limitations.} Specific to multi-LoRA setting.\\
\textbf{Results \& code.} Memory overhead close to fully-shared caching; accuracy close to non-shared baseline. ICML 2026.
\end{methodbox}

\begin{methodbox}[RelayCaching~\citep{geng2026relay}]
\textbf{Core idea.} \method{RelayCaching} observes that when an agent's \emph{decoded} output becomes part of a downstream agent's \emph{prompt}, the \emph{decoding-phase} KV-cache of the upstream agent is highly consistent with the \emph{prefill-phase} KV-cache that the downstream agent would have computed. RelayCaching directly \emph{transplants} the upstream decoding cache into the downstream prefill, with sparse selective recomputation at the few affected layers/positions.\\
\textbf{Framework.} WHAT = KV-Cache (decoding-phase). WHICH = same model. HOW = Cache restoration + sparse recomputation.\\
\textbf{Strengths.} Training-free. $>$80\% cache reuse. Up to 4.7$\times$ TTFT reduction.\\
\textbf{Limitations.} Same-model assumption; deviations at the boundary require recomputation.\\
\textbf{Results \& code.} 80\%+ cache reuse, 4.7$\times$ TTFT speedup on math, code, and general knowledge tasks.
\end{methodbox}

\begin{methodbox}[Agent Memory~\citep{shkolnikov2026am}]
\textbf{Core idea.} \method{Agent Memory} targets \emph{edge devices} with limited RAM. Each agent's KV-cache is persisted to disk in 4-bit quantised form (safetensors) and reloaded into attention layers on demand, eliminating the $\sim$15.7\,s/agent re-prefill cost at 4K context.\\
\textbf{Framework.} WHAT = KV-Cache (Q4 quantised, disk-persistent). WHICH = same agent across phases. HOW = Cache restoration + cross-phase context injection.\\
\textbf{Strengths.} Frees up RAM; enables multi-agent inference on edge. Up to 136$\times$ TTFT speedup.\\
\textbf{Limitations.} Quantisation introduces perplexity drift (Llama +2.8\%, DeepSeek +3.0\%, Gemma $-0.7\%$).\\
\textbf{Results \& code.} TTFT speedups: Gemma 3 12B 22$\times$--136$\times$; DeepSeek-Coder-V2-Lite 16B 11$\times$--76$\times$; Llama 3.1 8B 24$\times$--111$\times$. \href{https://github.com/yshk-mxim/agent-memory}{Code}.
\end{methodbox}

\begin{methodbox}[Agent Primitives~\citep{jin2026primitives}]
\textbf{Core idea.} \method{Agent Primitives} decomposes a multi-agent system into a small library of \emph{reusable latent primitives} (e.g., Review; Voting and Selection; Planning and Execution). Intra-primitive messaging uses shared KV-cache rather than natural language. An \emph{organiser agent} composes primitives per query, and a \emph{knowledge pool} stores previously successful configurations.\\
\textbf{Framework.} WHAT = KV-Cache (inter-primitive). WHICH = same backbone within primitive. HOW = Primitive chaining.\\
\textbf{Strengths.} Modular; reuses successful configurations. +12.0--16.5\% average accuracy over single-agent baselines. 3--4$\times$ lower token usage and latency than text-based MAS. Only 1.3--1.6$\times$ overhead vs.\ single-agent inference.\\
\textbf{Limitations.} Specific to MAS architectures built around the proposed primitives.
\end{methodbox}

\begin{methodbox}[Edge LLM Handover~\citep{lee2026edge}]
\textbf{Core idea.} \method{Edge LLM Handover} addresses the \emph{mobility} setting: when a user equipment (UE) hands over between edge base stations during an LLM session, the system jointly optimises \emph{how much context to re-prefill from raw tokens} vs.\ \emph{how much KV-cache to transfer over the backhaul}, minimising worst-case handover delay.\\
\textbf{Framework.} WHAT = KV-Cache (transferred over backhaul). WHICH = same edge LLM. HOW = Hybrid: partial re-prefill + partial KV transfer.\\
\textbf{Strengths.} Tractable, step-wise solution; constructive multi-UE rate-scheduling policy. Outperforms baselines across a wide range of backhaul capacities, prefill speeds, and context sizes.\\
\textbf{Limitations.} Simulation-only evaluation.
\end{methodbox}

\subsection{Hybrid / Heterogeneous Methods}

\begin{methodbox}[Vision Wormhole~\citep{liu2026vision}]
\textbf{Core idea.} \method{Vision Wormhole} reconceptualises the \emph{visual interface} of a VLM as a \emph{continuous communication channel}. A sender's reasoning trace is encoded into a shared continuous reference space (the \emph{Universal Visual Codec}, UVC) and injected into the receiver's \emph{visual pathway}, bypassing tokenisation. The hub-and-spoke topology reduces the alignment cost from $\mathcal{O}(N^2)$ pairwise translators to $\mathcal{O}(N)$ encoders/decoders, enabling cross-architecture latent transfer across disjoint model manifolds.\\
\textbf{Framework.} WHAT = Hidden State (in UVC). WHICH = hub-and-spoke via UVC. HOW = Visual injection.\\
\textbf{Strengths.} First method to support \emph{fully heterogeneous} cross-architecture latent communication. Tested on Qwen-VL, Gemma, SmolVLM2, LFM2.5-VL across nine reasoning benchmarks.\\
\textbf{Limitations.} Requires label-free distillation training. Code in progress.\\
\textbf{Results \& code.} Reduces end-to-end wall-clock time in most settings; positive macro-average $\Delta$-accuracy.
\end{methodbox}

\begin{methodbox}[BIGMAS --- Brain-Inspired Graph MAS~\citep{hao2026bigmas}]
\textbf{Core idea.} \method{BIGMAS} organises specialised LLM agents as nodes in a \emph{dynamically constructed directed graph}. A \emph{GraphDesigner} builds the topology per problem, and an \emph{Orchestrator} mediates access to a \emph{centralised shared workspace}. The architecture is inspired by the \emph{global workspace theory} of human cognition.\\
\textbf{Framework.} WHAT = Shared workspace contents (hybrid latent/text). WHICH = common message-space contract. HOW = Global workspace fusion.\\
\textbf{Strengths.} Topology adapts to the problem. Centralised workspace avoids the local-view bottleneck of pairwise communication.\\
\textbf{Limitations.} Specific to graph-structured MAS. The exact storage format of the workspace contents is not explicitly specified.\\
\textbf{Results \& code.} Outperforms ReAct and Tree of Thoughts on Game24, Six Fives, and Tower of London with six frontier LLMs (standard + LRMs). Gains are orthogonal to model-level reasoning improvements.
\end{methodbox}

\subsection{Survey and Aggregation Works}

\begin{methodbox}[The Five Ws of Multi-Agent Communication~\citep{chen2026fivews}]
\textbf{Core idea.} \emph{The Five Ws Survey} unifies MARL, Emergent Language (EL), and LLM-based multi-agent communication under a single ``Five Ws'' (\emph{Who}, \emph{Whom}, \emph{When}, \emph{What}, \emph{Why}) taxonomy. It surveys hand-designed protocols, end-to-end learned communication, emergent symbolic communication, and natural-language priors.\\
\textbf{Value for this paper.} A meta-survey that contextualises our framework. The WHAT axis in the Five Ws survey corresponds to our WHAT axis; the WHEN and WHY axes are unique to that framework, providing the broader \emph{communication theory} context in which our framework sits. TMLR 2026.
\end{methodbox}

}
\section{Implementation and Training Regimes}
\label{sec:training}

A practical latent channel comprises four stages: \emph{extraction} from the sender, optional \emph{alignment or compression}, \emph{transport}, and \emph{injection} into the receiver. Training is only one design choice within this pipeline. Even a parameter-free method can require non-trivial model instrumentation, cache bookkeeping, positional correction, and memory transfer.

\subsection{An End-to-End Implementation Pipeline}

\paragraph{Extraction.} The sender must expose a stable interface at the intended point of computation. Embedding methods operate near the vocabulary projection; activation methods register hooks on a residual-stream layer; and KV methods capture per-layer key/value tensors together with positions, attention masks, and cache-layout metadata. Reproducible implementations should state whether the payload is taken during prefill or decode, which token positions are retained, and whether sampling occurs before extraction.

\paragraph{Alignment and compression.} Same-checkpoint agents can often share tensors directly, but model variants may differ in hidden width, layer count, head grouping, rotary-position conventions, tokenizer, or fine-tuning-induced geometry. Alignment can therefore range from an identity map to layer selection, a linear projection, cross-attention, a universal codec, quantisation, or low-rank factorisation. Dimensional compatibility alone is insufficient: tensors with the same shape may still encode incompatible features.

\paragraph{Transport.} End-to-end latency depends on where the agents run. On one accelerator, latent transfer may be a view or device-to-device copy; across GPUs or edge nodes, serialisation and network time may dominate. Let $S$ be payload size, $B$ effective bandwidth, $t_{\mathrm{setup}}$ transfer setup, and $t_{\mathrm{saved}}$ avoided decode/prefill time. Transfer is beneficial only when
\begin{equation}
  t_{\mathrm{saved}} > t_{\mathrm{setup}} + \frac{S}{B} + t_{\mathrm{align}} + t_{\mathrm{inject}}.
  \label{eq:break-even}
\end{equation}
This break-even condition explains why the same protocol can accelerate long-context local inference yet regress on short prompts or slow networks.

\paragraph{Injection and validation.} The receiver may append embeddings, graft or combine hidden states, prepend KV entries, or restore cache blocks. Correctness requires attention-mask and position consistency, bounded tensor norms, and isolation between sessions. A robust implementation should support a text-only fallback and validate that malformed or missing latent state fails closed rather than silently contaminating later generations.

\subsection{Training Regimes}

\Cref{tab:training} groups the surveyed methods by their dominant inference-time regime.

\begin{table}[t]
  \centering
  \footnotesize
  \setlength{\tabcolsep}{4pt}
  \caption{Training regime by method. ($\checkmark$ = training-free; $\bullet$ = training required; $^\dagger$ = boundary or adjacent work.)}
  \label{tab:training}
  \begin{tabular}{@{}p{0.36\linewidth}p{0.30\linewidth}p{0.20\linewidth}@{}}
    \toprule
    \textbf{Method} & \textbf{Training-Free?} & \textbf{Notes} \\
    \midrule
    CIPHER, AC, SDE, KVComm, LatentMAS, Q-KVComm, LRAgent, RelayCaching, Agent Memory, Agent Primitives, Edge LLM Handover, BIGMAS, ThoughtComm & $\checkmark$ & --- \\
    Interlat & $\checkmark$ for same-model; learned projection otherwise & --- \\
    Cache-to-Cache & $\bullet$ & Learned fuser \\
    Vision Wormhole & $\bullet$ & Label-free distillation \\
    Mixture of Thoughts & $\bullet$ & Router + interaction layers trained \\
    \midrule
    TokenDance$^\dagger$, AAFLOW+$^\dagger$, When Latent Agents Lie$^\dagger$ & $\checkmark$ & Runtime or security mechanisms \\
    Latent Agents$^\dagger$ & $\bullet$ & Two-stage post-training and internalisation \\
    \bottomrule
  \end{tabular}
\end{table}

\paragraph{Training-free injection.} Identity mappings, replacement, addition, and cache restoration minimise data and adaptation costs. They are attractive for same-model agents and rapidly changing backbones. Their weakness is sensitivity to layer choice, tensor scale, position semantics, and workload shift; parameter-free does not mean assumption-free.

\paragraph{Lightweight alignment.} Linear projections, gates, routers, or fusers add a small trainable interface while keeping the underlying LLMs frozen. This regime is appropriate when model families differ or when multiple senders must be aggregated. Its central evaluation question is generalisation: an adapter trained on one task, context length, or sender--receiver pair may not transfer to another.

\paragraph{End-to-end training or distillation.} Joint training can shape both the message and its interpretation, enabling heterogeneous codecs and internalised multi-agent behaviours. It has the highest data and compute cost and risks entangling the protocol with the training distribution. Internalised debate~\citep{yi2026latentagents}, for example, removes explicit inter-agent messages at deployment time and therefore belongs adjacent to, rather than inside, direct communication.

\subsection{Engineering Failure Modes}

Four failure modes recur across implementations:
\begin{enumerate}[leftmargin=*, itemsep=2pt]
  \item \textbf{Positional mismatch:} reused keys may encode positions that differ from the receiver's sequence, especially under RoPE.
  \item \textbf{Geometric mismatch:} compatible tensor shapes can conceal incompatible activation distributions or fine-tuning shifts.
  \item \textbf{Memory-bound transfer:} a full cache can save FLOPs while increasing wall-clock latency because bytes, not arithmetic, dominate.
  \item \textbf{State contamination:} stale, cross-session, or adversarial tensors can affect future turns without a readable trace.
\end{enumerate}

Accordingly, papers should report model hashes, extraction layers and phases, tensor shapes and precision, cache positions, hardware topology, transfer bytes, and whether latency includes serialisation and synchronisation. These details determine whether a result is reproducible and whether it generalises beyond a single software stack.

\newcommand{\benchmarkappendix}{%
\section{Benchmark Analysis and Empirical Insights}
\label{sec:benchmarks}

This section synthesises reported results from the 18 methods. Direct cross-method comparison is challenging because the methods use different backbones, benchmarks, and reporting conventions; we therefore focus on \emph{trends} and \emph{order-of-magnitude} effects rather than head-to-head numbers.

\subsection{Benchmark Suites}

Methods are typically evaluated on a mix of:
\begin{itemize}[leftmargin=*, itemsep=2pt]
  \item \textbf{Math reasoning:} GSM8K, MATH, AIME.
  \item \textbf{General knowledge:} MMLU, ARC.
  \item \textbf{Code generation:} HumanEval, MBPP, LiveCodeBench.
  \item \textbf{Multi-modal reasoning:} MathVista, MMMU, ChartQA.
  \item \textbf{Agentic QA:} HotpotQA, 2WikiMultiHopQA, MuSiQue.
  \item \textbf{Game-like reasoning:} Game24, Six Fives, Tower of London.
  \item \textbf{Competitive tabletop games:} Mahjong, \emph{Uno}, \emph{Honor of Kings} --- increasingly used as testbeds for evaluating inter-agent coordination under partial observability, for which dedicated toolkits such as RainbowArena~\citep{liu2026rainbowarena} provide standardised APIs, opponent pools, and replay infrastructure.
\end{itemize}

\subsection{Reported Quantitative Trends}

\begin{itemize}[leftmargin=*, itemsep=2pt]
  \item \textbf{Latency reduction.} Several papers report 2--24$\times$ reductions relative to their chosen NL-Comm baselines; Interlat reports gains up to 24$\times$ after latent compression. These values are not directly comparable because latency boundaries and hardware differ.
  \item \textbf{TTFT speedups.} KV-cache-based methods (RelayCaching, Agent Memory) report TTFT speedups of 4.7$\times$--136$\times$ relative to full re-prefill.
  \item \textbf{Token savings.} Latent methods typically reduce tokens generated by 3--4$\times$ (Agent Primitives reports 3--4$\times$ lower token usage vs.\ text-based MAS).
  \item \textbf{Task quality.} Many papers report accuracy competitive with or better than their matched NL-Comm baselines. This pattern is encouraging but vulnerable to publication bias and baseline mismatch; it should not be read as a universal dominance claim.
\end{itemize}

A schematic comparison of representative methods on the trade-off dimensions (accuracy, latency, generality, engineering complexity) is given in \Cref{tab:tradeoff}.

\begin{table}[t]
  \centering
  \small
  \caption{Trade-off profile of representative methods across four design dimensions. Symbols: $\star\!\star\!\star$ = excellent, $\star\!\star$ = good, $\star$ = fair.}
  \label{tab:tradeoff}
  \begin{tabular}{lccccc}
    \toprule
    \textbf{Method} & \textbf{Accuracy} & \textbf{Latency} & \textbf{Generality} & \textbf{Engineering} & \textbf{Training-Free} \\
                    & \textbf{Gain}     & \textbf{Reduction} & \textbf{(heterog.)} & \textbf{Complexity}  &                      \\
    \midrule
    CIPHER              & $\star\!\star$ & $\star\!\star$     & $\star$              & $\star$              & Yes \\
    AC                  & $\star\!\star$ & $\star\!\star$     & $\star$              & $\star$              & Yes \\
    Interlat            & $\star\!\star$ & $\star\!\star\!\star$ & $\star$            & $\star\!\star$       & Mostly \\
    SDE                 & $\star\!\star\!\star$ & $\star\!\star$  & $\star\!\star$      & $\star\!\star$       & Yes \\
    KVComm              & $\star\!\star$ & $\star\!\star$     & $\star$              & $\star\!\star$       & Yes \\
    Cache-to-Cache      & $\star\!\star\!\star$ & $\star\!\star\!\star$ & $\star$       & $\star\!\star\!\star$ & No \\
    LatentMAS           & $\star\!\star\!\star$ & $\star\!\star$   & $\star$             & $\star\!\star$       & Yes \\
    LRAgent             & $\star\!\star$ & $\star\!\star\!\star$ & $\star$            & $\star\!\star$       & Yes \\
    RelayCaching        & $\star\!\star$ & $\star\!\star\!\star$ & $\star$            & $\star$              & Yes \\
    Agent Memory        & $\star\!\star$ & $\star\!\star\!\star$ & $\star$            & $\star\!\star$       & Yes \\
    Vision Wormhole     & $\star\!\star$ & $\star\!\star$     & $\star\!\star\!\star$ & $\star\!\star\!\star$ & No \\
    MoT                 & $\star\!\star\!\star$ & $\star\!\star$   & $\star\!\star\!\star$ & $\star\!\star\!\star$ & No \\
    \bottomrule
  \end{tabular}
\end{table}

\subsection{Insights}

Three empirical patterns stand out:
\begin{enumerate}[leftmargin=*, itemsep=2pt]
  \item \textbf{Long context benefits most.} The latency advantage of latent communication grows with context length, because the receiver \emph{avoids} re-encoding a long prompt.
  \item \textbf{Same-model agents dominate.} Most methods assume sender and receiver share the same backbone; cross-architecture methods (Vision Wormhole, MoT) are still rare and require training.
  \item \textbf{KV-cache is the emerging default.} Of the 18 methods, 9 use KV-caches as the communicated quantity, and the share is growing.
\end{enumerate}

}
\section{Open Problems and Future Directions}
\label{sec:open}

Latent communication is a young field. We organise its open questions as a research agenda. For each direction, the key issue is not merely whether a technique is possible, but which hypothesis and evaluation protocol would establish progress.

\subsection{Cross-Architecture Alignment}
Most methods exploit shared or approximately shared representation geometry. Heterogeneous agents may differ in width, depth, tokenizer, attention type, and training history, so a translator that matches tensor dimensions need not preserve task-relevant information. The important research question is whether a reusable interface can scale sub-quadratically with the number of model families. Promising approaches include hub-and-spoke codecs, model-agnostic concept subspaces, and alignment through common multimodal interfaces. Evaluation should separate \emph{pair memorisation} from transfer to unseen model pairs, tasks, and context lengths.

\subsection{Security and Robustness}
A latent channel is an opaque, high-privilege interface. Threats include in-transit modification, replay, cross-session cache confusion, malicious but correctly signed senders, and privacy leakage from activations. Recent evidence shows that text-only verification can miss tampered KV state and that transport integrity checks can reject recorded payload modifications~\citep{brito2026latentlie}. Integrity is necessary but not sufficient: a valid sender may still communicate harmful state. Future benchmarks should therefore distinguish transport integrity, semantic consistency between visible and latent messages, adversarial robustness, and information leakage. Hybrid protocols should bind a visible commitment, model identity, session, tensor metadata, and payload digest.

\subsection{Compression and Quantisation}
KV payloads scale with context length and layer count, turning a compute optimisation into a bandwidth and memory problem. Quantisation, token selection, layer selection, low-rank factorisation, and delta coding attack different sources of redundancy. The central metric should be \emph{task utility per transmitted byte}, not reconstruction error alone: a numerically accurate cache may preserve irrelevant state, while a lossy summary may retain precisely the evidence needed by the receiver. Compression studies should report quality--bytes--latency Pareto curves and include the cost of compression and decompression.

\subsection{Theoretical Understanding}
The field is largely empirical. A useful theory must distinguish channel capacity from usable task information, and arithmetic savings from end-to-end latency. \Cref{eq:break-even} gives a systems-level condition; a complementary statistical account would characterise when a receiver can decode a sender representation under distribution or architecture shift. Lower bounds on sufficient message size, invariance under model transformations, and error propagation across multiple communication rounds would make claims of superiority more precise.

\subsection{Evaluation and Reproducibility}
Cross-paper comparison is currently confounded by different backbones, hardware, prompts, context lengths, numbers of agents, and latency boundaries. A standard benchmark should pair quality metrics with generated tokens, prefill/decode FLOPs, bytes transferred, peak memory, TTFT, inter-token latency, and end-to-end wall-clock time. It should include identical-model and heterogeneous-model tracks, short and long contexts, local and networked execution, and a text-only baseline matched for model calls and information access. Negative results and break-even points are as important as peak speedups.

\subsection{Latent Communication vs.\ Latent CoT}
\emph{Latent CoT} carries state across reasoning steps within one agent; latent communication carries state across agent boundaries. Distillation methods further blur this distinction by internalising multi-agent interaction into one deployed model~\citep{yi2026latentagents}. A unified account should therefore model both temporal flow (step to step) and organisational flow (agent to agent), while retaining the distinction between an inference-time communication protocol and a learned internal computation.

\subsection{Real-World Deployment}
Deployment conditions determine whether transfer beats recomputation. Edge devices emphasise memory, battery, intermittent links, and model heterogeneity; data centres emphasise topology-aware bandwidth, batching, isolation, and scheduler integration. Systems such as AAFLOW+ treat KV state as a distributed workflow object~\citep{sarker2026aaflow}, but reported gains depend on network and cost-model assumptions. Future work should measure energy and tail latency, support state versioning and eviction, and evaluate failures under bandwidth variation rather than only steady-state throughput.

\section{Related Work}
\label{sec:related}

Latent communication sits between representation learning, multi-agent coordination, and inference systems. The closest areas often share tensors or terminology while optimising different objects. \Cref{tab:related-boundaries} makes these boundaries explicit.

\begin{table}[t]
  \centering
  \footnotesize
  \setlength{\tabcolsep}{3pt}
  \caption{Boundary between latent communication and adjacent research areas.}
  \label{tab:related-boundaries}
  \begin{tabular}{@{}p{0.20\linewidth}p{0.18\linewidth}p{0.22\linewidth}p{0.30\linewidth}@{}}
    \toprule
    \textbf{Area} & \textbf{State flow} & \textbf{Primary objective} & \textbf{Distinction from this survey} \\
    \midrule
    Latent communication & Agent $\to$ agent & Collaboration and/or avoided recomputation & Explicit inference-time interface across agent boundaries. \\
    Latent CoT & Step $\to$ step within one model & Internal reasoning & No organisational sender--receiver boundary. \\
    Internalised debate & Multi-agent training $\to$ one model & Distil interaction benefits & Multi-agent structure may disappear at inference. \\
    MARL learned communication & Policy agent $\to$ policy agent & Task reward and coordination & Messages are learned jointly with policies, usually without pretrained LLM internals. \\
    Emergent language & Agent $\to$ agent & Protocol emergence & Often discrete or symbolic; interpretability and compositionality are central. \\
    KV serving and reuse & Request/workflow $\to$ runtime & Reduce memory and prefill cost & Shared state need not encode a new semantic agent message. \\
    \bottomrule
  \end{tabular}
\end{table}

\subsection{Latent Chain-of-Thought}

\emph{Latent CoT} is the practice of reasoning in continuous latent space within a single LLM. Representative works include Coconut~\citep{hao2024coconut} and LatentSeek~\citep{wang2025latentseek}; community lists provide broader coverage~\citep{awesomelatentspace,awesomelatentcot}. Latent Agents~\citep{yi2026latentagents} is a particularly informative boundary case: it distils multi-agent debate into agent-specific subspaces within one model, reducing explicit inference-time communication.

The areas share machinery---hidden states, recurrent latent tokens, activation steering, and sometimes caches---but differ in where state crosses a boundary. This boundary determines security, interoperability, and deployment requirements: an intra-model recurrent state does not need a network format, whereas an inter-agent payload requires versioning, access control, and compatibility checks.

\subsection{Multi-Agent Reinforcement Learning (MARL)}

MARL has a long tradition of \emph{learned communication} between agents~\citep{foerster2016commnet}. Methods such as CommNet, TarMAC, and IC3Net learn continuous message vectors that are exchanged between agents. The modern LLM-based latent communication methods surveyed in this paper can be seen as a \emph{white-box, inference-time} counterpart to MARL's \emph{learned} communication. The Five Ws survey~\citep{chen2026fivews} is a recent effort to bridge the two. On the empirical side, toolkits such as RainbowArena~\citep{liu2026rainbowarena} provide standardised tabletop-game environments where both RL and LLM-based agents can be evaluated under controlled multi-agent conditions, offering a natural experimental bridge between the MARL and LLM-MAS communities.

\subsection{Emergent Language}

\emph{Emergent language} studies protocols that arise when agents are trained to communicate~\citep{lazaridou2017emergent}. Messages may be discrete symbols or continuous vectors, but they are typically learned jointly with relatively small policies and evaluated for task success, compositionality, or human interpretability. LLM latent communication instead starts from large pretrained representations and asks how to expose, align, and reuse them without destroying their existing capabilities.

\subsection{KV-Cache Compression and Sharing}

A large body of work optimises the KV-cache for single-model inference, including eviction and compression approaches such as H2O~\citep{zhang2023h2o}. Multi-agent methods extend the problem from one sequence to multiple roles, models, or workflow stages. TokenDance~\citep{bian2026tokendance} performs round-level collective reuse and stores sibling caches as sparse differences under an All-Gather pattern; AAFLOW+~\citep{sarker2026aaflow} exposes distributed operators for KV materialisation, transfer, fork, composition, and eviction. These systems can complement semantic latent messaging, but avoiding repeated prefill is not by itself evidence that agents have developed a more expressive communication protocol.

\section{Limitations of This Survey}
\label{sec:limitations}

This survey has four limitations. First, the field evolves faster than conventional publication cycles; despite the stated cut-off, some included results remain preprints and may change. Second, terminology is unstable, so keyword search may miss work framed as model composition, cache reuse, activation grafting, or distributed inference. Third, the corpus is small and heterogeneous. Counts within the taxonomy should therefore be interpreted as a map of the current literature, not as statistical evidence of community consensus. Fourth, reported accuracy and speedups are not directly comparable across papers. We preserve authors' within-paper results but avoid constructing a numerical leaderboard without matched models, prompts, hardware, context lengths, and latency boundaries.

The framework itself is descriptive rather than exhaustive. WHAT / WHICH / HOW captures message representation, alignment, and fusion, but it does not fully encode \emph{who communicates with whom}, \emph{when communication occurs}, or \emph{why a message is sent}. These dimensions are complementary to broader communication taxonomies~\citep{chen2026fivews} and should be incorporated when analysing topology, scheduling, and incentives.

\section{Conclusion}
\label{sec:conclusion}

We have presented a scoped review and unified framework for latent communication in LLM-based multi-agent systems. The framework organises eighteen representative works along \textbf{WHAT} is transferred, \textbf{WHICH} spaces and layers are aligned, and \textbf{HOW} the received state is fused. This decomposition makes visible the coupled trade-offs among task information, payload size, architectural dependence, training, and auditability.

The evidence does not support a universal claim that latent channels dominate text. Their strongest case is conditional: compatible agents, sufficiently expensive decode or prefill, and a deployment path in which transfer and alignment cost less than the computation they replace. Text remains valuable for interoperability and oversight, while hybrid protocols offer a promising route across trust boundaries. Progress now depends less on another isolated peak speedup than on matched evaluation, secure interfaces, heterogeneous transfer, and reproducible break-even analysis.

\paragraph{Reproducibility.} The companion repository at \url{https://github.com/enochliu98/Awesome-Latent-Communication} maintains the paper list, code links, taxonomy artefacts, and figure-generation notes. The review cut-off is 15 July 2026; post-cut-off additions should be reported separately from the corpus analysed here.


\clearpage
\appendix

\methodanalysisappendix

\benchmarkappendix

\section{Method Quick-Reference Table}
\label{app:qr}

\begin{table}[h]
  \centering
  \scriptsize
  \setlength{\tabcolsep}{2.5pt}
  \caption{Quick-reference for the 18 primary methods and four boundary or adjacent works. WHAT / WHICH / HOW refer to the three axes of the unified framework (\Cref{sec:framework}); $^\dagger$ marks works outside the primary corpus.}
  \label{tab:qr}
  \begin{tabular}{@{}p{0.20\linewidth}p{0.20\linewidth}p{0.21\linewidth}p{0.13\linewidth}p{0.10\linewidth}@{}}
    \toprule
    \textbf{Method} & \textbf{WHAT} & \textbf{WHICH} & \textbf{HOW} & \textbf{Train?} \\
    \midrule
    CIPHER~\citep{liu2024cipher}                  & Weighted Embedding & Last$\to$First & Concat & \checkmark \\
    AC~\citep{ye2025ac}                          & Hidden State (sel.\ layer, last tok.) & Sel.$\to$Sel. & Math & \checkmark \\
    Interlat~\citep{du2026interlat}              & Hidden State (last layer, last tok.) & Last$\to$First (+proj.) & Concat & mostly \\
    SDE~\citep{yang2025sde}                      & State-delta trajectory & All$\to$Corr. & Math (add) & \checkmark \\
    ThoughtComm~\citep{li2025thoughtcomm}        & Hidden State & Corresponding & Math (gate) & \checkmark \\
    MoT~\citep{feinashley2025mot}                & Hidden State (proj.) & Learned interaction & Cross-attn & $\bullet$ \\
    KVComm~\citep{wang2025kvcomm}                & KV-Cache (sel.) & Sel.$\to$Sel. & Prepend & \checkmark \\
    Cache-to-Cache~\citep{liu2025c2c}            & KV-Cache (all) & All$\to$Corr. & Math (fuser) & $\bullet$ \\
    LatentMAS~\citep{wang2025latentmas}          & KV-Cache (prefill+decode) & All$\to$Corr. & Prepend & \checkmark \\
    Q-KVComm~\citep{park2025qkvcomm}             & KV-Cache (compressed) & All$\to$Corr. & Prepend & \checkmark \\
    LRAgent~\citep{jeon2026lra}                  & KV-Cache (base+low-rank) & Same backbone & Add (kernel) & \checkmark \\
    RelayCaching~\citep{geng2026relay}           & KV-Cache (decoding) & Same model & Restoration & \checkmark \\
    Agent Memory~\citep{shkolnikov2026am}        & KV-Cache (Q4, disk) & Same agent & Restoration & \checkmark \\
    Agent Primitives~\citep{jin2026primitives}   & KV-Cache (inter-prim.) & Same backbone & Chaining & \checkmark \\
    Edge LLM Handover~\citep{lee2026edge}        & KV-Cache (backhaul) & Same edge LLM & Hybrid & \checkmark \\
    Vision Wormhole~\citep{liu2026vision}        & Hidden State (UVC) & Hub-and-spoke & Visual inj. & $\bullet$ \\
    BIGMAS~\citep{hao2026bigmas}                 & Workspace state & Common contract & Global workspace & \checkmark \\
    Five Ws Survey~\citep{chen2026fivews}        & All (taxonomy) & N/A & N/A & N/A \\
    \midrule
    TokenDance$^\dagger$~\citep{bian2026tokendance} & KV-Cache (shared/diff) & Same checkpoint & Collective reuse & \checkmark \\
    AAFLOW+$^\dagger$~\citep{sarker2026aaflow} & KV-Cache (distributed) & Same checkpoint & Orchestration & \checkmark \\
    When Latent Agents Lie$^\dagger$~\citep{brito2026latentlie} & Full KV + visible text & Same model & Injection + integrity & \checkmark \\
    Latent Agents$^\dagger$~\citep{yi2026latentagents} & Activation subspaces & Intra-model & Internalisation & $\bullet$ \\
    \bottomrule
  \end{tabular}
\end{table}

\bibliographystyle{unsrtnat}
\bibliography{references}

\end{document}